\documentclass[10pt,twocolumn,letterpaper]{article}

\usepackage{iccv}
\usepackage{times}
\usepackage{epsfig}
\usepackage{graphicx}
\usepackage{amsmath}
\usepackage{amssymb}

\usepackage[pagebackref=true,breaklinks=true,letterpaper=true,colorlinks,bookmarks=false]{hyperref}
\usepackage[skip=0pt plus0pt, indent=10pt]{parskip}

\iccvfinalcopy 

\def\iccvPaperID{50} 

\ificcvfinal\fi

\begin{document}

\title{Geodesic Regression Characterizes \\ 3D Shape Changes in the Female Brain During Menstruation}

\author{Adele Myers\\
UC Santa Barbara\\
{\tt\small adele@ucsb.edu}
\and
Caitlin Taylor\\
UC Santa Barbara\\
{\tt\small cmtaylor@ucsb.edu}
\and
Emily Jacobs\\
UC Santa Barbara\\
{\tt\small emily.jacobs@gmail.com}
\and
Nina Miolane\\
UC Santa Barbara\\
{\tt\small ninamiolane@ucsb.edu}
}

\maketitle
\ificcvfinal\thispagestyle{empty}\fi

\begin{abstract}
    Women are at higher risk of Alzheimer's and other neurological diseases after menopause, and yet research connecting female brain health to sex hormone fluctuations is limited. We seek to investigate this connection by developing tools that quantify 3D shape changes that occur in the brain during sex hormone fluctuations. Geodesic regression on the space of 3D discrete surfaces offers a principled way to characterize the evolution of a brain's shape. However, in its current form, this approach is too computationally expensive for practical use. In this paper, we propose approximation schemes that accelerate geodesic regression on shape spaces of 3D discrete surfaces. We also provide rules of thumb for when each approximation can be used. We test our approach on synthetic data to quantify the speed-accuracy trade-off of these approximations and show that practitioners can expect very significant speed-up while only sacrificing little accuracy. Finally, we apply the method to real brain shape data and produce the first characterization of how the female hippocampus changes shape during the menstrual cycle as a function of progesterone: a characterization made (practically) possible by our approximation schemes. Our work paves the way for comprehensive, practical shape analyses in the fields of bio-medicine and computer vision. Our implementation is  \href{https://github.com/bioshape-lab/my28brains}{publicly available on GitHub}.
\end{abstract}

\section{Introduction}

Women are more likely to experience Alzheimer's, cognitive decline, and navigational issues after menopause~\cite{beam2018differences,vetvik2017sex}. Yet, topics relevant to female brain health such as menstruation, pregnancy, menopause, and their associated female sex hormone fluctuations only account for 0.3\% of the neuroimaging literature between 1995 and 2022~\cite{taylor2021scientific}. The hippocampal formation (a brain structure) is an excellent diagnostic tool for investigating the connection between female brain health and sex hormones, as it is the first cortical region to harbor neuropathology in the progression to Alzheimer's (causing characteristic shape changes visible on magnetic resonance images (MRI))~\cite{du2001magnetic,maleki2013common}, and it is also very sensitive to sex hormone fluctuations~\cite{galea2013sex}. Sex hormone fluctuations significantly influence brain anatomy and function in healthy subjects~\cite{taylor2020progesterone,taxier2020oestradiol,pritschet2020functional}. Enhancing our understanding of how hormonal fluctuations affect the healthy brain is crucial to explaining why females are later more at risk for neurological conditions after menopause~\cite{clayton2014policy}. We seek to close this knowledge gap by starting with one question: How does the hippocampal formation respond to monthly fluctuations in ovarian hormones during the menstrual cycle?

\begin{figure}[h!]
    \centering
    \includegraphics[width=1.\linewidth]{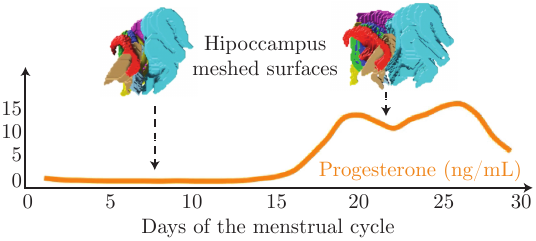}
    \caption{During the menstrual cycle, the ovaries cyclically release female sex hormones such as progesterone into the blood. We propose practical shape analysis tools that quantify 3D shape changes in the brain that occur during progesterone fluctuations, with a focus on the hippocampus: a structure involved in memory and navigation. Visualization created from data by \cite{taylor2020progesterone,pritschet2020functional}.}
    \label{fig:visualization}
\end{figure}

Recent research has shown that certain substructures of the hippocampal formation change their volume over the course of the menstrual cycle in response to progesterone, but no significant volumetric change was found on the whole-formation level~\cite{taylor2020progesterone}. Our 3D visualizations (Fig.~\ref{fig:visualization}) show that the hippocampal formation does change its \textit{shape} on a whole-formation level, but no team has quantified how this effect depends on progesterone levels. This is not surprising because quantifying 3D shape changes is technically challenging, and is in fact is an active research area in itself in mathematics and computer vision. For example, quantifying surface shape changes as a function of a continuous variable like progesterone can theoretically be performed through geodesic regression on Riemannian manifolds~\cite{fletcher2011geodesic,thomas2013geodesic}: an extension of linear regression dedicated to shape spaces. However, in its current form, geodesic regression on surface spaces is too slow for practical use. Here, we bridge the gap between computer vision and clinical neuroimaging by presenting a new practical method, a hybrid between geodesic and linear regression, that allows us to quantify how the \textit{shape} of the hippocampal formation changes in response to progesterone.

\paragraph{Contributions} We offer several contributions that span the fields of machine learning, differential geometry and clinical neuroimaging. First, in machine learning, we introduce our hybrid geodesic-linear regression method: a faster geodesic regression model which uses linear residuals instead of geodesic residuals in its loss function and also uses a linear regression result to initialize its geodesic regression optimization. Then, we perform extensive synthetic experiments to offer rules of thumb for deciding between linear, geodesic, and geodesic-linear regression. In differential geometry, these results give novel intuition about the curvature of the nonlinear data space of surface shapes. In clinical neuroimaging, these rules of thumb provide practitioners with guidelines to decide on the speed-accuracy trade-off between the regression types, revealing whether a surface mesh sequence can be adequately characterized by the considerably faster linear or geodesic-linear regressions without sacrificing accuracy. Finally, we apply our paradigm to real brain magnetic resonance images (MRIs) of a female brain through the menstrual cycle. We characterize, for the first time, shape changes in the female hippocampal formation as a function of progesterone.

\section{Related Works}

Consider a series of hippocampal shapes, with the surface of each shape described as a mesh extracted via segmentation from a full-brain MRI (Fig.~\ref{fig:visualization}). The \textit{shapes} of these discrete surfaces (meshes) can be described either \textit{extrinsically} or \textit{intrinsically}. The extrinsic approach with the Large Deformation Diffeomorphic Metric Mapping framework~\cite{beg2005computing} represents a surface shape in 3D space by the amount of deformation that one needs to apply on the whole 3D ambient grid containing the surface in order to deform a reference shape into the surface shape of interest. By contrast, the intrinsic approach only deforms the surface itself~\cite{bauer2021numerical}, and hence provides us with two advantages: (i) intuitive deformations that can be discussed with neuroscientists, (ii) higher computational efficiency with up to 10x acceleration~\cite{bauer2021numerical}. We focus here on the intrinsic approach.

\paragraph{Analysis of Parameterized Surfaces}
In the intrinsic approach, surfaces can be either \textit{parameterized} or \textit{unparameterized}. Each mesh in a dataset of parameterized surfaces is constrained to have a consistent structure: the vertices of meshes in the dataset have one-to-one correspondences. By contrast, datasets of unparameterized surfaces relax this constraint such that statistical analyses can be performed independently of the number and indexation of the mesh vertices. While theoretically grounded, the computational complexity of this approach makes it unpractical for geodesic regression. As an example, statistical analyses within this framework are often limited to population average computations and machine learning algorithms that rely only on distances such as multidimensional scaling or k-means clustering \cite{bauer2021numerical,hartman2023elastic,kurtek2010novel}.

Therefore, we consider the scenario where each mesh in the dataset is first parameterized to match a reference parameterization, and then statistical analysis such as regression is performed. In this scenario, the first natural choice is to consider linear regression on the set of 3D coordinates of the vertices. Linear regression has the advantage of being conceptually simple while enjoying analytical solutions: it will be our first baseline. However, it has the drawback that it does not enjoy parameterization-invariance. In other words, the distance (or dissimilarity) between two surfaces may change if we choose another reference parameterization, which may change the regression results. In the context of clinical application where the ultimate goal is human health, it is thought that we cannot afford such inconsistency. The alternative is to consider parameterization-invariant distances between parameterized surfaces. In differential geometry, this can be achieved by equipping the space of parameterized surfaces with a Riemannian metric that is invariant under reparameterizations~\cite{jermyn2017elastic,hartman2023elastic}. This process however, turns the data space into a nonlinear manifold, where linear regression needs to be generalized to geodesic regression~\cite{fletcher2011geodesic,thomas2013geodesic}. Geodesic regression does not enjoy a closed-form estimator and is typically computationally expensive. It will be our second baseline.

\begin{figure*}[htbp!]
    \centering
    \includegraphics[width=\textwidth]{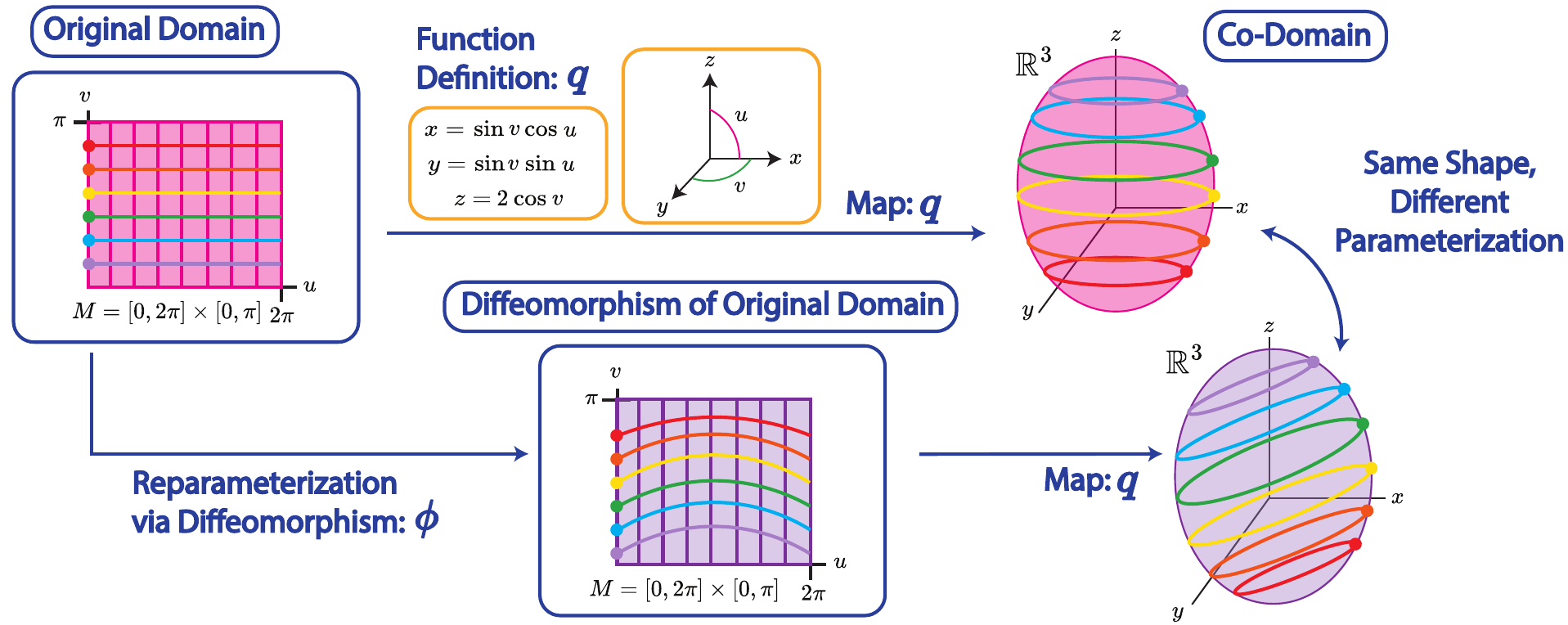}
    \caption{A surface is represented by a function $q: M \to \mathbb{R}^3$ that maps parameters $(u, v)\in M$ to points in 3D space $q(u, v) \in \mathbb{R}^3$ (top row). Its parameterization can be changed by applying a diffeomorphism $\phi$ to the domain $M$ before mapping to $\mathbb{R}^3$ (bottom row).}\label{fig:continuous_surface_reparameterized}
\end{figure*}

\paragraph{Computational Challenges of Geodesic Regression}
Geodesic regression~\cite{fletcher2011geodesic,thomas2013geodesic} for parameterization-invariant Riemannian metrics presents unique computational challenges. In the Riemannian framework, calculating a single geodesic requires a computationally expensive approach. Geodesic regression solves an optimization problem by minimizing a mean square error (MSE) loss function. The MSE requires the computations of $n+1$ geodesics at each iteration: 1 geodesic representing the generative model, and $n$ geodesics required to compute the residuals in the MSE, where $n$ is the number of surfaces (meshes) in the dataset. We observe that works developing the Riemannian framework limit the number of geodesic computations required for their analysis, and do not perform any form of regression. Kilian et al. \cite{kilian2007geometric} focus on geodesic interpolation or extrapolation of parameterized surfaces and do not study regression. Kurtek et al. \cite{kurtek2010novel} limit their experimental analysis to computing geodesics between pairs of unparameterized surfaces and performing clustering. Jermyn et al. \cite{jermyn2017elastic} compute geodesics between pairs of parameterized surfaces and provide a classification experiment. Hartman et al \cite{hartman2023elastic} estimate population averages and perform dimension reduction with multidimensional scaling (MDS) and tangent PCA for parameterized and unparameterized surfaces. Bauer et al \cite{bauer2021numerical} compute geodesics and the population averages of unparameterized surfaces, together with multi-dimensional scaling and k-means clustering. We suspect that the authors did not perform geodesic regression in these works because of their computational costs, which we investigate here.

\section{Background}\label{sec:background}

This section presents the mathematical background necessary to formulate our approximation schemes for geodesic regression on the shape space of (hippocampal) surfaces. We refer to \cite{hartman2023elastic,guigui2023introduction} for additional details.

\paragraph{A. Riemannian Metrics and Geodesics} We first introduce concepts in differential geometry necessary for geodesic regression. A \textit{Riemannian metric} on a smooth manifold $\mathcal{N}$ is a family $\left(G_p\right)_{p \in \mathcal{N}}$ of inner products on each tangent space $T_p \mathcal{N}$, such that $G_p$ depends smoothly on the point $p\in \mathcal{N}$. Any Riemannian metric $G$ yields a notion of distance between points $q_0, q_1$ on $\mathcal{N}$. Specifically, if $\gamma : [0,1] \rightarrow \mathcal{N}$ is a smooth trajectory on $\mathcal{N}$ with velocity vector at $t \in [0, 1]$ denoted as $\dot{\gamma}_{t} \in T_{\gamma(t)}\mathcal{N}$, its length is defined as $L_{\gamma} = \int_{0}^1 \sqrt{G(\dot{\gamma}_{t}, \dot{\gamma}_{t})_{\gamma_t}} dt$ and the distance between any two points $q_0, q_1 \in \mathcal{N}$ is given by $d(q_0, q_1) = \inf_{\gamma : \gamma(0)=q_0, \gamma(1)=q_1} L_{\gamma}$.

A \textit{geodesic} between two points $q_0, q_1$ that are ``close'' in $\mathcal{N}$ is defined as a trajectory $\gamma$ that locally realizes the shortest distance between $q_0$ and $q_1$. Intuitively, a geodesic is the generalization to manifolds of the concept of  a straight line in vector spaces. While some manifolds enjoy analytical expression for their geodesics, this is not case for the manifold of (hippocampal) surface shapes that we will consider here. Thus, geodesics will need to be computed numerically. 

To this aim, geodesics are expressed as the solutions of the \textit{geodesic equation}, which is an ordinary differential equation (ODE) which can be written in local coordinates as:
\begin{equation}\label{eq:geodesic_ode}
    \ddot{\gamma}^k(t)+\Gamma_{i j}^k \dot{\gamma}^i(t) \dot{\gamma}^j(t)=0,
\end{equation}
for all times $t \in[0, 1]$ where $\Gamma_{i j}^k$ are the Christoffel symbols associated with the Riemannian metric. Solving this ODE provides numerical solutions for geodesics. 

To perform geodesic regression, we will also need two additional operations, called Exp and Log, which we define here. The map $(q, v) \mapsto \gamma_{q, v}(1)$ defined for $(q, v) \in \mathcal{I} \times T_q\mathcal{I}$ is called the \textit{exponential map} (Exp) and essentially computes the point $\gamma_{q, v}(1)$ after following the geodesic of initial point $q \in \mathcal{N}$ and initial velocity $v \in T_q\mathcal{N}$. The inverse of the Exp map on its injectivity domain is called the \textit{logarithm map} (Log).

\paragraph{B. Surfaces and Their Parameterizations} A \textit{continuous surface} can be described by a function $q: M \to \mathbb{R}^3$, where $M$ is a two-dimensional space of parameters $(u, v) \in M$ that parameterize the 3D points $q(u, v) \in \mathbb{R}^3$ on the surface. Intuitively, the function $q$ deforms the space of parameters $M$ to give the surface its distinct shape, e.g., the ellipsoid shown in the top row of Fig.~\ref{fig:continuous_surface_reparameterized}. Mathematically, $q$ is required to be an oriented smooth mapping in $C^\infty(M, \mathbb{R}^3)$ that is also regular in the sense that its differential $dq$ is injective everywhere on $M$.

The \textit{parameterization of a surface} refers to the placement of points on the surface. If we define one surface as $q: M \to \mathbb{R}^3$, then we can describe the same surface with a different parameterization by $q \circ \phi: M \to \mathbb{R}^3$, where $\phi$ is an orientation-preserving diffeomorphism of $M$. Intuitively, $\phi$ smoothly deforms the placement of parameters on the domain $M$, which in turn smoothly changes the placement of points in the co-domain, as shown with rainbow colors in the bottom row of Fig.~\ref{fig:continuous_surface_reparameterized}. The change of parameterization $\phi$ does not change the shape of the surface, which is an ellipsoid in both rows of Fig.~\ref{fig:continuous_surface_reparameterized}.

\paragraph{C. Space of Surfaces} The space of surfaces is denoted $\mathcal{I} \subset C^\infty(M, \mathbb{R}^3)$. The space $\mathcal{I}$ is an infinite dimensional manifold immersed in the infinite dimensional vector space $C^{\infty}(M, \mathbb{R}^3)$. The Riemannian metric we choose to equip the manifold $\mathcal{I}$ with defines the distance between its points $q_0, q_1 \in \mathcal{I}$ and thus the notion of dissimilarity between the two surfaces $q_0, q_1$. We consider the second-order Sobolev metric~\cite{hartman2023elastic}:
\begin{equation} \label{eq:Sobolev}
\begin{aligned}
G_q(h, k) & =\int_M\left(a_0\langle h, k\rangle+a_1 g_q^{-1}\left(d h_m, d k_m\right)\right. \\
& +b_1 g_q^{-1}\left(d h_{+}, d k_{+}\right)+c_1 g_q^{-1}\left(d h_{\perp}, d k_{\perp}\right) \\
& \left.+d_1 g_q^{-1}\left(d h_0, d k_0\right)+a_2\left\langle\Delta_q h, \Delta_q k\right\rangle\right) \operatorname{vol}_q,
\end{aligned}
\end{equation}
where $h, k$ are tangent vectors at point $q \in \mathcal{N}$; $g_q^{-1}$ is the pullback metric from $\mathbb{R}^3$ that defines distances on the surface $q$ itself; $\Delta_q$ is the Laplacian induced by $q$; $dh_m, dh_+,dh_\perp, dh_0$ are orthogonal vector-valued one-forms and $\text{vol}_q$ is the surface area measure of $q$. The scalars $a_0, a_1, a_2, b_1, c_1, d_1$ are weighting parameters that define distance between two surfaces based on how they are sheared, scaled, bent, or parameterized with respect to each other. 

The choice of second-order Sobolev metric is motivated by the following facts. First, the zero-order and first-order Sobolev metrics yield less stable results in geodesic interpolation between complex 3D shapes ~\cite{hartman2023elastic}. Second, the weighting parameters $a_0, a_1, a_2, b_1, c_1, d_1$ defining the second-order Sobolev metric in Eq.~\eqref{eq:Sobolev} can be linked to observable physical deformations (shearing, bending, etc) which helps with intuitively comparing physical objects. Last, the metric in Eq.~\eqref{eq:Sobolev} yields a distance that is rotation and reparameterization invariant~\cite{hartman2023elastic}. In other words, if all the surfaces in the dataset are rotated and reparameterized in the same way, i.e., using the same rotation matrix and reparameterization diffeomorphism $\phi$, then their pairwise distances are unchanged. We note that this property is practical only if we first assume that all the surfaces (are oriented and) have valid point-to-point correspondences.

\begin{figure}[h!]
    \centering
    \includegraphics[width=\linewidth]{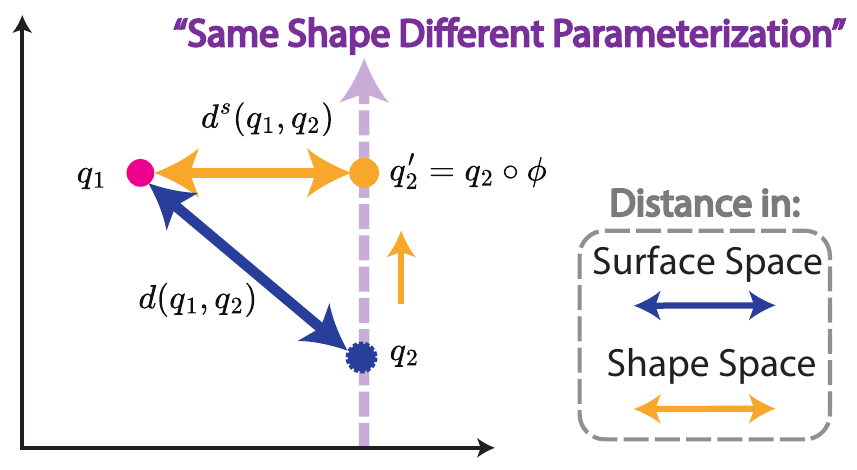}
    \caption{Distances in surface space vs shape space. $q_1$ and $q_2$ are two surfaces with different parameterization and different shape. The distance given by the second-order Sobolev metric \ref{eq:Sobolev} measures both the parameterization and shape differences. The shape space distance \ref{eq:shape-distance} only measures difference in shape.}
    \label{fig:shape_space}
\end{figure}

\paragraph{D. Space of Surface Shapes} In the space of surfaces $\mathcal{I}$, if two surfaces have the same shape but different orientations or parameterizations, they correspond to different points. By contrast, we introduce the space of surface shapes where two surfaces with the same shape correspond to the same point, regardless of differences in their orientation or parameterization. Mathematically, the space of surface shapes is defined as the quotient space: $\mathcal{I} / (\text{Rot}(\mathbb{R}^3) \times \text{Diff}(M))$ \textemdash see \cite{hartman2023elastic} for details. For simplicity, we consider the case of parameterizations with the shape space $\mathcal{S} = \mathcal{I} / \text{Diff}(M)$ while the case of orientations can be treated similarly. 

In the shape space $\mathcal{S}$, the distance between two surface shapes $q_1$ and $q_2$ is given by:
\begin{equation} \label{eq:shape-distance}
    d^\mathcal{S}(q_1, q_2)= \text{inf}_{\phi} d(q_1, q_2 \circ \phi)
    \\ = d(q_1, q_2'),
\end{equation}
where $\phi$ represents a choice in parameterization. In Eq.~\eqref{eq:shape-distance}, the parameterization of $q_2$ is varied until the second-order Sobolev distance $d$ in Eq.~\eqref{eq:Sobolev} between $q_1$ and $q_2$ reaches an infimum as shown in Fig.~\ref{fig:shape_space}. This operation matches the parameterization of $q_2$ to the parameterization of $q_1$ so that any remaining discrepancy between them is due to difference in shape, rather than difference in parameterization. Ideally, we would perform our geodesic regression methods directly in the shape space $\mathcal{S}$. However, the high computational cost of this approach leads us to instead compute in the surface space $\mathcal{I}$ after choosing a reference parameterization that corresponds to the first hippocampal surface of our dataset.

\section{Methods}
\begin{figure*}[h!]
    \centering
    \includegraphics[width=\textwidth]{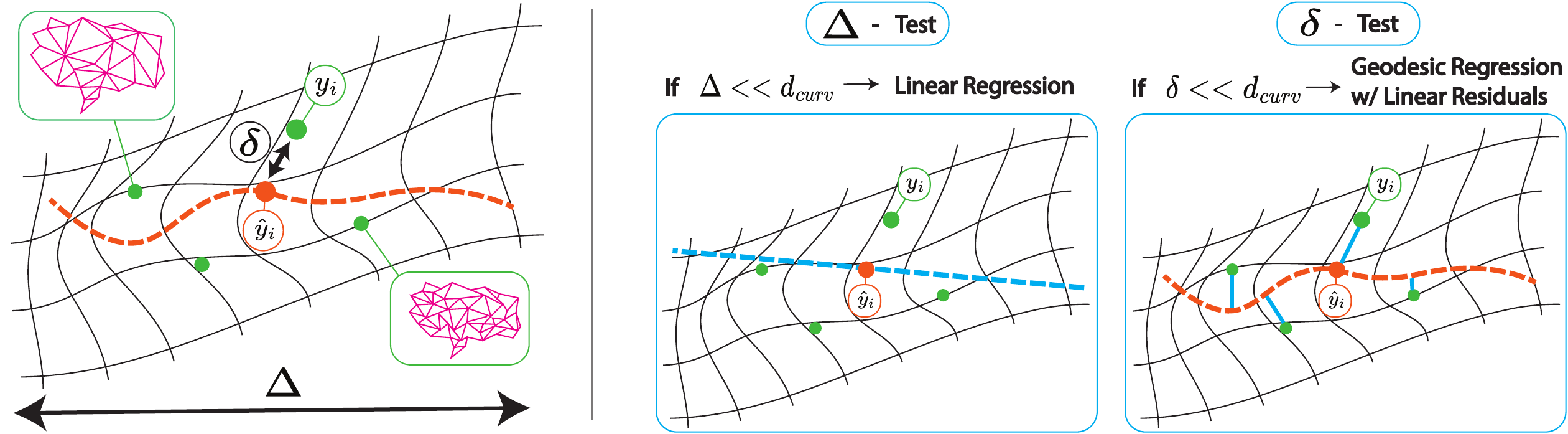}
    \caption{Overview: Approximation schemes for geodesic regression approximation.  $\delta$-test: if the residual magnitudes are small compared to the curvature of the manifold, we can use geodesic regression with linear residuals (GRLR). $\Delta$-test: if the distance covered by the data set is small compared to the curvature, we use linear regression (LR). Reducing geodesic regression (GR) to either GRLR or LR provides up to four orders of magnitude speed-up, while sacrificing little accuracy.}
    \label{fig:overview}
\end{figure*}

We seek to quantify the \textit{anatomical changes} in the hippocampal formation that emerge from progesterone variations during the menstrual cycle. To achieve this, we propose approximations to geodesic regression on the space of 3D brain shapes that make it computationally fast enough for practical use. We further propose rules of thumb for determining when each approximation can be used, as summarized in Fig.~\ref{fig:overview}. 

\subsection{Linear Regression}
\paragraph{Model} \textit{Linear regression} (LR) models the relationship between an independent variable $X \in \mathbb{R}$ and the dependent variable $Y$ taking values in $\mathbb{R}^D$ as:
\begin{equation}\label{eq:linear_model}
    Y=\alpha+X \beta+\epsilon,
\end{equation}
where $\alpha \in \mathbb{R}^D$ is the intercept, $\beta \in \mathbb{R}^D$ is the slope, and $\epsilon$ represents the noise.

\paragraph{Loss} Given data $\left(x_i, y_i\right) \in$ $\mathbb{R} \times \mathbb{R}^D$, for $i=1, \ldots, n$, we fit the linear regression model through least squares, i.e., we compute the estimates for the intercept and slope $\hat{\alpha}, \hat{\beta}$ as:
\begin{equation}\label{eq:linear_leastsq}
    (\hat{\alpha}, \hat{\beta})=\arg \min _{(\alpha, \beta)} \frac{1}{2}\sum_{i=1}^n\left\|y_i-\hat{y}_i\right\|^2
    \text{ for } \hat{y}_i = \alpha + x_i \beta,
\end{equation}
which minimizes the summed squared magnitude of the (linear) \textit{residuals}:  $y_i-\alpha-x_i \beta$, for $i=1, ..., n$.

\paragraph{Learning} Importantly for computational purposes, this minimization has an analytical solution given by the normal equations $\hat{\beta}=\frac{\frac{1}{n} \sum x_i y_i-\bar{x} \bar{y}}{\sum x_i^2-\bar{x}^2}$ and $\hat{\alpha}=\bar{y}-\bar{x} \hat{\beta}$,
where $\bar{x}$ and $\bar{y}$ are the sample means of the $x_i$ and $y_i$, respectively. We will use linear regression as our first baseline, where $X$ is the level of progesterone, and $Y$ is the hippocampal surface discretized as a mesh, which takes values in $\mathbb{R}^{N \times 3}$ where $N$ is the number of mesh vertices.

\subsection{Geodesic Regression}
\paragraph{Model} \textit{Geodesic regression} (GR)~\cite{fletcher2011geodesic,thomas2013geodesic} models the relationship between an independent variable $X \in \mathbb{R}$ and the dependent variable $Y$, whose values lie on a manifold $\mathcal{N}$, as:
\begin{equation}\label{eq:geodesic_model} 
Y=\operatorname{Exp}(\operatorname{Exp}(p, X v), \epsilon),
\end{equation}
where $\epsilon$ is noise in the tangent space at $\operatorname{Exp}(p, X v)$, and $\text{Exp}$ is the operation defined in the previous section. Note that when the manifold of interest is $\mathcal{N}=\mathbb{R}^D$, the exponential operator simplifies to addition: $\operatorname{Exp}(p, v)=$ $p+v$. Consequently, the geodesic regression generative model simplifies to the linear regression generative model of Eq.~\eqref{eq:linear_model} with $p=\alpha$ and $v = \beta$. We also note that the exponential operation appears twice: to model the geodesic itself, and to model the noise $\epsilon$. In what follows, we consider geodesic regression on the manifold $\mathcal{N} = \mathcal{I}$ equipped with a second-order Sobolev metric from Eq.~\eqref{eq:Sobolev}.

\paragraph{Loss} Given data $\left(x_i, y_i\right) \in \mathbb{R} \times \mathcal{I}$, for $i=1, \ldots, n$, we seek to learn estimates of the intercept and slope $(p, v) \in \mathcal{I} \times T_p\mathcal{I}$. In the manifold setting, the loss function associated with the geodesic given by $(p, v)$ is minimized as:
\begin{align}\label{eq:geodesic_leastsq}
    (\hat{p}, \hat{v})
        &= \arg \min _{(p, v)} \frac{1}{2} \sum_{i=1}^n d\left(y_i, \hat{y}_i\right)^2, \\
        &= \arg \min _{(p, v)} \frac{1}{2} \sum_{i=1}^n \|\text{Log}(\hat{y}_i, y_i)\|_{\hat{y}_i}^2, \\
        &\text{for } \hat{y}_i = \operatorname{Exp}\left(p, x_i v\right).
\end{align}
We compute estimates for intercept and slope  $(\hat{p}, \hat{v})$ which minimize the summed squared magnitude of the (geodesic) \textit{residuals} $\text{Log}(\operatorname{Exp}\left(p, x_i v\right), y_i)$ for $i=1, ..., n$. The geodesic residuals differ from the linear residuals as they are calculated with exponentials and logarithms instead of additions and subtractions.  

\paragraph{Learning} In contrast to linear regression, the least squares problem of Eq.~\eqref{eq:geodesic_leastsq} above does not have an analytical solution for general manifolds $\mathcal{I}$. Instead, we need to compute the estimates of the intercept and slope with gradient descent, which is typically computationally expensive. Gradient descent comes in two flavors depending on the strategy used to compute the gradient, which can be either a Riemannian gradient as originally proposed in ~\cite{fletcher2011geodesic} or an extrinsic gradient. The Riemannian gradient writes~\cite{fletcher2011geodesic}:
$$
\begin{aligned}
& \nabla_p l=-\sum_{i=1}^N d_p \operatorname{Exp}\left(p, x_i v\right)^{\dagger} \epsilon_i, \\
& \nabla_v l=-\sum_{i=1}^N x_i d_v \operatorname{Exp}\left(p, x_i v\right)^{\dagger} \epsilon_i,
\end{aligned}
$$
where $l$ is the loss function, $\epsilon_i=\text{Log} \left(\operatorname{Exp}\left(p, x_i v\right), y_i\right)$ are the residuals, $d_v$ and $d_p$ are derivatives, and $\dagger$ denotes the adjoint. In the general case, the expression of these derivatives and their respective adjoint operators are not known although they can be derived analytically for some manifolds as in \cite{thomas2013geodesic}. However, to the best of our knowledge, no such formula exists for shape spaces of parameterized surfaces, so we use a numerical approach.

\subsection{Why is Geodesic Regression Slow?}

The geodesic regression optimization is slow due to the Exp and Log maps in Eq.~\eqref{eq:geodesic_leastsq}. Computation of the exponential and logarithm maps do not enjoy an analytical expression for the manifold that we are interested in, and neither do their differentials. Consequently, we compute them only numerically, as implemented in Geomstats~\cite{Miolane2020geomstats} as follows.

For the \textit{exponential map}, we consider the geodesic equation as a coupled system of first-order ODEs:
$$
\left\{\begin{array}{l}
v(t)=\dot{\gamma}(t) \\
\dot{v}(t)=f(v(t), \gamma(t), t)
\end{array}\right.
$$
where $f$ is a smooth function given by Eq.~\eqref{eq:geodesic_ode} and the state variable is $(\gamma(t), \dot{\gamma}(t))$. Given initial conditions, we use a first-order forward Euler scheme to integrate this system. For a given step $dt$ we compute:
\begin{equation}
    v(t+dt) = v(t) + \dot{v}(t) dt = v(t) + f(v(t), \gamma(t), t) dt.
\end{equation}
Introducing the parameter $n_{steps}$, if we integrate this geodesic equation between $t=0$ and $t=1$ in $n_{\text{steps}}$ then we use $dt = (n_{\text{steps}})^{-1}$. Consequently, the parameter $n_{\text{steps}}$ controls the numerical precision of the computation of the exponential map. The computation of Exp is slow due to this numerical integration.

For the \textit{logarithm map}, we solve the optimization problem in $v$:
$$
\min d^2\left(\operatorname{Exp}(p, v), q\right),
$$
that represents the fact that Log is the inverse map of Exp. This minimization is solved by gradient descent (GD) until a convergence tolerance is reached. It uses scipy for minimization method and computes the gradient of the exponential map with automatic differentiation. The computation of Log is slow due to this optimization process.

\subsection{Approximations Schemes with Rules of Thumb}

Curved spaces are locally linear. Thus, if a data set falls on a ``small'' portion the shape space, addition and subtraction will offer excellent approximations of exponentials and logarithms and avoid costly computations. To speed up geodesic regression, we propose two approximation schemes shown in Fig.~\ref{fig:overview}: (i) linear regression, and (ii) geodesic regression with linear residuals, where (ii) represents a novel approach for geometric machine learning\textemdash which we describe in the next section.

We also propose rules of thumb to determine when each of these methods will yield sufficiently accurate approximations of geodesic regression.  For (i), we propose the $\Delta$-Test, which explores when the magnitude of the geodesic length of the data set ($\Delta$) is small at the scale of the curvature of the manifold (Fig.~\ref{fig:overview} middle). For (ii), we propose the $\delta$-Test, which explores when the magnitude of the noise ($\delta$) is small at the scale of the curvature of the manifold (Fig.~\ref{fig:overview} right). In the experiments section, we explore the curvature of the space of 3D discrete surfaces to give numerical values to these guidelines.

\subsection{Geodesic Regression with Linear Residuals}

\paragraph{Model} We propose \textit{geodesic regression with linear residuals} (GRLR) to model the relationship between an independent variable $X \in \mathbb{R}$, the noise-free dependent variable taking values in a manifold, and the (noisy) dependent variable $Y$ taking values in $\mathbb{R}^D$. In other words, we propose the following generative model:
\begin{equation}\label{eq:geodesiclinear_model}
Y = \operatorname{Exp}(p, X v) + \epsilon,
\end{equation}
where $\operatorname{Exp}(p, X v)$ is the noise-free dependent variable and $\epsilon$ is the noise. The noise-free dependent variable is constrained to be a surface in $\mathcal{I}$, and thus $Y$'s dependency on $X$ is modelled using the Exp operation. However, in practical applications, the data's noise may push the data off of $\mathcal{I}$. Thus, in addition to its computational gain, this generative model acknowledges the fact that there is no reason for the noise to be constrained on the manifold.

\paragraph{Loss} Given data $\left(x_i, y_i\right) \in \mathbb{R} \times \mathbb{R}^D$, for $i=1, \ldots, n$, we fit this regression model through least squares, i.e., we compute the estimates for the intercept and slope as:
\begin{equation}\label{eq:geodesiclinear_leastsq}
(\hat{p}, \hat{v})=\arg \min _{(p, v)} \frac{1}{2} \sum_{i=1}^n \|\hat{y}_i - y_i\|^2\text{ for } \hat{y}_i =
\operatorname{Exp}\left(p, x_i v\right),
\end{equation}
where the squared \textit{geodesic distance} of Eq.~\ref{eq:geodesic_leastsq} has been replaced by the \textit{squared Euclidean distance}, but we keep the exponential map defining the geodesic. 

\paragraph{Learning} Like geodesic regression, this least squares problem still requires gradient descent. The gradient can be computed as a Riemannian gradient or as an extrinsic gradient. The Riemannian gradient is given by:
$$
\begin{aligned}
& \nabla_p l=-\sum_{i=1}^N d_p \operatorname{Exp}\left(p, x_i v\right)^{\dagger} \epsilon_i, \\
& \nabla_v l=-\sum_{i=1}^N x_i d_v \operatorname{Exp}\left(p, x_i v\right)^{\dagger} \epsilon_i,
\end{aligned}
$$
where $l$ is the loss function, $\epsilon_i= y_i -\operatorname{Exp}\left(p, x_i v\right)$ are the residuals,  $d_v$ and $d_p$ are derivatives, and $\dagger$ is the adjoint. By avoiding the computation of $n$ logarithms, the learning process enjoys a significant speed-up. Even within the use of the extrinsic gradient, the loss function avoids the computations of these logarithms and is thus accelerated. We note that we still need to compute the Exp by numerical integrations and their derivatives which we do by automatic differentiation. Our implementation is  \href{https://github.com/bioshape-lab/my28brains}{publicly available on GitHub}.

\section{Experiments}
We investigate the curvature of the space of surfaces to quantify which approximation scheme should be used on which dataset. Guided by this analysis, we approximate geodesic regression on 3D hippocampal surfaces, giving the first characterization of hippocampal formation's shape change as a function of progesterone.

\subsection{Curvature Estimation with $\delta-$ and $\Delta-$ Tests}

\paragraph{Simulations} We perform experiments on synthetic meshes to provide rules-of-thumb that help the practitioner decide when linear regression or geodesic regression with linear residuals can be used with little loss in accuracy on the space of discrete surfaces.

Specifically, our experiments explore cases when lines can be used to approximate geodesics. First, we compute both a line and a geodesic between two meshes $q^{\text{start}}$ and $q^{\text{end}}$.  Then, we compare the meshes along the line in $\mathbb{R}^{N\times 3}$ to the meshes along the geodesic in $\mathcal{I}$, where $N$ is the number of vertices in the 3D meshes. We have either $n=5$ or $n=10$ meshes along each sequence. The start mesh $q^{\text{start}}$ is an ellipsoid whose principal axes have length 2, 2, and 3. The end mesh is a deformed version $q^{\text{end}}$ of the reference $q^{\text{start}}$, where the amount of deformation is controlled by a factor that we vary in $\{1\%, 10\%, 50\%, 100\%\}$. This deformation factor indicates by how much each vertex in $q^{\text{end}}$ has been moved compared to the vertex's position in $q^{\text{start}}$ and is given as a percentage of the diameter of $q^{\text{start}}$. $q^{end}$ is generated by adding isotropic Gaussian noise to each vertex in $q^{start}$. In other words, the standard deviation of the Gaussian noise is: $\sigma = \text{deformation} \times D$ where $D$ is the diameter of the mesh.

We determine how much the geodesic and the line between $q^{\text{start}}$ and $q^{\text{end}}$ differ by computing the root mean square deviation (RMSD) between the geodesic mesh sequence and the line mesh sequence, which we then normalize by the diameter $D$ of the mesh:
\begin{equation}
    \text{RMSD} = \frac{1}{D} \sqrt{\frac{1}{TN} \sum_{t = 1}^T \sum_{j = 1}^N \| v_{tj}^{\text{line}} - v_{tj}^{\text{geodesic}}\|^2}, 
\end{equation}
where $N$ is the number of vertices, $T$ the number of meshes in the sequence (5 or 10) and $D$ the diameter. We also time the computation of the geodesic and the line and report their ratio.

\begin{figure}[h!]
    \centering
    \includegraphics[width=\linewidth]{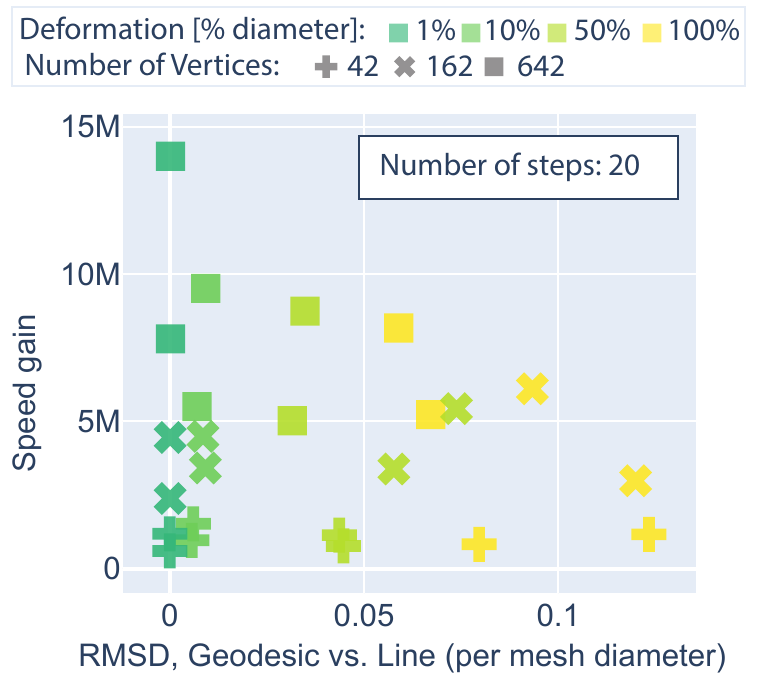}
    \caption{Accuracy-speed trade-off between a geodesic and its linear approximation in joining two meshes $q^{\text{start}}$ and $q^{\text{end}}$. The x-axis (error) shows how their meshes differ by computing distances between their vertices. The y-axis (speed) shows the ratio of their computational times. The color represents the deformation factor, i.e. how deformed the mesh $q^{\text{end}}$ is from the reference mesh $q^{\text{start}}$. The symbols represent the number of vertices in the meshes. The number of steps is a parameter controlling the numerical integration computing the geodesic. The results are similar when $n_{steps}$ = 5.}
    \label{fig:line_vs_geodesic}
\end{figure}

\begin{figure*}[h!]
    \centering
    \includegraphics[width=\linewidth]{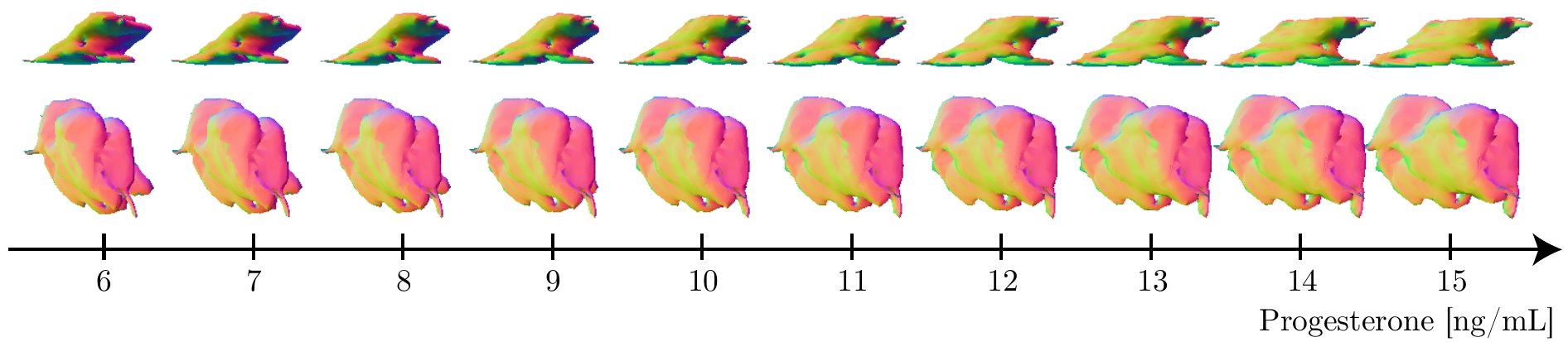}
    \caption{Linear regression reveals hippocampal deformation associated with an increase in progesterone during the menstrual cycle. The two rows show two views of each 3D mesh along the sequence. For visualization purposes, a coloring of the mesh is used to represent depth. While volumetric analyses did not capture a volumetric change, our shape analysis reveals that an increase in progesterone corresponds to a shear deformation of the hippocampal formation.}
    \label{fig:linear_regression_real_data}
\end{figure*}

\paragraph{$\delta$-Test} Fig.~\ref{fig:line_vs_geodesic} shows that deformation factors of 1\% yield errors below 0.05\% of the diameter of the shape (below 0.0005 on the figure). This is true across two values of the number of steps used for the numerical computation of the exponential map: $n_{steps}$ = 20, and $n_{steps}$ = 5  (see Section~\label{sec:background}). Consequently, for our $\delta$-Test: when the measurement noise on the vertices of the meshed shapes is expected to be less than 1\% of the total mesh diameter, we recommend using linear residuals instead of geodesic residuals. In this case, we assess that the shape manifold can be approximated as linear at the scale of residual length: the curvature is low compared to the magnitude of the noise. By using linear residuals, we enjoy a considerable speed-up, up to 14M $\times$ (for a number of steps of 20), as shown in Fig.~\ref{fig:line_vs_geodesic}, and up to 1.5M $\times$ (for a number of steps of 5) .

As a real-world example, consider $q^{\text{start}}$ as the mesh corresponding to a true hippocampal shape at a given level of progesterone, and $q^{\text{end}}$ as the mesh that we observe in practice after segmenting and extracting the mesh from the MRI data. In this case, the ``measurement noise'' is the MRI noise and the segmentation error. We expect measurement noise to displace each vertex by around 1\% of the total mesh diameter, since MRI images have a good resolution and both segmentation and meshing algorithms are reasonably accurate. Thus, brain MRIs are in the regime where geodesic regression can enjoy considerable speed-ups by utilizing linear residuals.

\paragraph{$\Delta$-Test} Additionally, Fig.~\ref{fig:line_vs_geodesic} shows that deformation factors of 10-50\% yield an error that is less than 10\% of the diameter (below 0.1 on the figure). Consequently, for our $\Delta$-Test: if the data set's largest deformation between two meshes is 10-50\% of the diameter of the mesh, and if practitioners can tolerate a maximum loss of accuracy of 10\% in their results, then using linear regression instead of geodesic regression allows them to significantly speed up their pipeline. In this case, we assess that the shape manifold can be approximated as linear on the scale of the data set: the curvature is low compared to the spread of the data. Even more strikingly, the error decreases from 10\% to only 3\% if practitioners consider meshes with hundreds of vertices, as shown in Fig.~\ref{fig:line_vs_geodesic}. 


\subsection{Hippocampal Shape Change Characterization}

\paragraph{Data set} We use a time-series of 3D brain images recorded from magnetic resonance imaging (MRI): 11 images from 11 consecutive days, capturing the progesterone peak of a single female subject's natural menstrual cycle~\cite{pritschet2020functional}, as analyzed by volumetric analyses in~\cite{taylor2020progesterone}. We choose to focus on the progesterone peak (11 days), as opposed to the full menstrual cycle (30 days) for simplicity. The female also measured hormone levels in her blood in conjunction with each MRI session.

\paragraph{Pre-processing} We align the 3D images to correct for the position and orientation of the subject's head in the MRI scanner, and to extract the surface of each substructure of the hippocampal formation. The results of this pre-processing are shown in Figure~\ref{fig:visualization} where each sub-structure surface is color-coded and shown at two different levels of progesterone (low and high). Here, we can visually observe the hippocampal formation's shape evolution, which our analyses will seek to characterize with the slope and intercept learned from the proposed approximation of geodesic regression. We then use Eq.~\eqref{eq:shape-distance} to give every hippocampal mesh the same parameterization. 

\paragraph{Characterization} Consider $q^{\text{start}}$, $q^{\text{end}}$ the meshes corresponding to hippocampus shapes at the lowest and highest progesterone levels respectively. Fig.~\ref{fig:visualization} shows that there is a deformation factor of around 10\%: each vertex is displaced by around 10\% of the total mesh diameter between the two meshes shown. Thus, following $\Delta$-Test, we use linear regression to provide a characterization of 3D shape changes in the hippocampal formation during the menstrual cycle. Fig.~\ref{fig:linear_regression_real_data}
reveals for the first time that the hippocampal formation shears in response to an increase in progesterone. Additionally, this computation provides an important educational tool. Clinical neuroscientists can use the result of our regression model to query, for a given progesterone level, what is the associated hippocampal shape.

\section{Conclusion}

We have proposed a shape analysis technique to reveal what volumetric analyses could not: that the overall shape of hippocampal formation changes during progesterone level fluctuation. The implications for women's health are profound. Because each structure of the brain is dedicated to a specific function, and the hippocampal formation is directly related to functions that deteriorate in women after menopause, characterizing how the hippocampal formation changes in response to sex hormones changes is critical. Not only does it provide a diagnostic for disease prediction, but it also offers a method to probe relationships between hormone level, hippocampal shape, and brain health.

Here, we provide a practical method to characterize such changes with slopes and intercepts learned through approximation schemes for geodesic regression on the space of 3D discrete surfaces. This work aims to open research avenues for automated, fast, and statistically sound diagnostics of female brain health.

{\small
\bibliographystyle{ieee_fullname}
\bibliography{egbib}

\begin{thebibliography}{10}\itemsep=-1pt

\bibitem{bauer2021numerical}
Martin Bauer, Nicolas Charon, Philipp Harms, and Hsi-Wei Hsieh.
\newblock A numerical framework for elastic surface matching, comparison, and
  interpolation.
\newblock {\em International Journal of Computer Vision}, 129(8):2425--2444,
  2021.

\bibitem{beam2018differences}
Christopher~R Beam, Cody Kaneshiro, Jung~Yun Jang, Chandra~A Reynolds, Nancy~L
  Pedersen, and Margaret Gatz.
\newblock Differences between women and men in incidence rates of dementia and
  alzheimer’s disease.
\newblock {\em Journal of Alzheimer's disease}, 64(4):1077--1083, 2018.

\bibitem{beg2005computing}
M~Faisal Beg, Michael~I Miller, Alain Trouv{\'e}, and Laurent Younes.
\newblock Computing large deformation metric mappings via geodesic flows of
  diffeomorphisms.
\newblock {\em International journal of computer vision}, 61:139--157, 2005.

\bibitem{clayton2014policy}
Janine~A Clayton and Francis~S Collins.
\newblock Policy: Nih to balance sex in cell and animal studies.
\newblock {\em Nature}, 509(7500):282--283, 2014.

\bibitem{du2001magnetic}
AT Du, Nea Schuff, D Amend, MP Laakso, YY Hsu, WJ Jagust, K Yaffe, JH Kramer, B
  Reed, D Norman, et~al.
\newblock Magnetic resonance imaging of the entorhinal cortex and hippocampus
  in mild cognitive impairment and alzheimer's disease.
\newblock {\em Journal of Neurology, Neurosurgery \& Psychiatry},
  71(4):441--447, 2001.

\bibitem{fletcher2011geodesic}
Thomas Fletcher.
\newblock Geodesic regression on riemannian manifolds.
\newblock In {\em Proceedings of the Third International Workshop on
  Mathematical Foundations of Computational Anatomy-Geometrical and Statistical
  Methods for Modelling Biological Shape Variability}, pages 75--86, 2011.

\bibitem{galea2013sex}
Liisa~AM Galea, Steven~R Wainwright, MM Roes, P Duarte-Guterman, C Chow, and DK
  Hamson.
\newblock Sex, hormones and neurogenesis in the hippocampus: hormonal
  modulation of neurogenesis and potential functional implications.
\newblock {\em Journal of neuroendocrinology}, 25(11):1039--1061, 2013.

\bibitem{guigui2023introduction}
Nicolas Guigui, Nina Miolane, Xavier Pennec, et~al.
\newblock Introduction to riemannian geometry and geometric statistics: from
  basic theory to implementation with geomstats.
\newblock {\em Foundations and Trends{\textregistered} in Machine Learning},
  16(3):329--493, 2023.

\bibitem{hartman2023elastic}
Emmanuel Hartman, Yashil Sukurdeep, Eric Klassen, Nicolas Charon, and Martin
  Bauer.
\newblock Elastic shape analysis of surfaces with second-order sobolev metrics:
  a comprehensive numerical framework.
\newblock {\em International Journal of Computer Vision}, 131(5):1183--1209,
  2023.

\bibitem{jermyn2017elastic}
Ian~H Jermyn, Sebastian Kurtek, Hamid Laga, Anuj Srivastava, Gerard Medioni,
  and Sven Dickinson.
\newblock {\em Elastic shape analysis of three-dimensional objects}.
\newblock Springer, 2017.

\bibitem{kilian2007geometric}
Martin Kilian, Niloy~J Mitra, and Helmut Pottmann.
\newblock Geometric modeling in shape space.
\newblock In {\em ACM SIGGRAPH 2007 papers}, pages 64--es. 2007.

\bibitem{kurtek2010novel}
Sebastian Kurtek, Eric Klassen, Zhaohua Ding, and Anuj Srivastava.
\newblock A novel riemannian framework for shape analysis of 3d objects.
\newblock In {\em 2010 IEEE computer society conference on computer vision and
  pattern recognition}, pages 1625--1632. IEEE, 2010.

\bibitem{maleki2013common}
Nasim Maleki, Lino Becerra, Jennifer Brawn, Bruce McEwen, Rami Burstein, and
  David Borsook.
\newblock Common hippocampal structural and functional changes in migraine.
\newblock {\em Brain Structure and Function}, 218:903--912, 2013.

\bibitem{Miolane2020geomstats}
Nina Miolane, Nicolas Guigui, Alice {Le Brigant}, Johan Mathe, Benjamin Hou,
  Yann Thanwerdas, Stefan Heyder, Olivier Peltre, Niklas Koep, Hadi Zaatiti,
  Hatem Hajri, Yann Cabanes, Thomas Gerald, Paul Chauchat, Christian Shewmake,
  Daniel Brooks, Bernhard Kainz, Claire Donnat, Susan Holmes, and Xavier
  Pennec.
\newblock {Geomstats: A Python package for Riemannian geometry in machine
  learning}.
\newblock {\em Journal of Machine Learning Research}, 21:1--9, 2020.

\bibitem{pritschet2020functional}
Laura Pritschet, Tyler Santander, Caitlin~M Taylor, Evan Layher, Shuying Yu,
  Michael~B Miller, Scott~T Grafton, and Emily~G Jacobs.
\newblock Functional reorganization of brain networks across the human
  menstrual cycle.
\newblock {\em Neuroimage}, 220:117091, 2020.

\bibitem{taxier2020oestradiol}
Lisa~R Taxier, Kellie~S Gross, and Karyn~M Frick.
\newblock Oestradiol as a neuromodulator of learning and memory.
\newblock {\em Nature Reviews Neuroscience}, 21(10):535--550, 2020.

\bibitem{taylor2021scientific}
Caitlin~M Taylor, Laura Pritschet, and Emily~G Jacobs.
\newblock The scientific body of knowledge--whose body does it serve? a
  spotlight on oral contraceptives and women’s health factors in
  neuroimaging.
\newblock {\em Frontiers in neuroendocrinology}, 60:100874, 2021.

\bibitem{taylor2020progesterone}
Caitlin~M Taylor, Laura Pritschet, Rosanna~K Olsen, Evan Layher, Tyler
  Santander, Scott~T Grafton, and Emily~G Jacobs.
\newblock Progesterone shapes medial temporal lobe volume across the human
  menstrual cycle.
\newblock {\em NeuroImage}, 220:117125, 2020.

\bibitem{thomas2013geodesic}
P Thomas~Fletcher.
\newblock Geodesic regression and the theory of least squares on riemannian
  manifolds.
\newblock {\em International journal of computer vision}, 105:171--185, 2013.

\bibitem{vetvik2017sex}
Kjersti~Gr{\o}tta Vetvik and E~Anne MacGregor.
\newblock Sex differences in the epidemiology, clinical features, and
  pathophysiology of migraine.
\newblock {\em The Lancet Neurology}, 16(1):76--87, 2017.

\end{thebibliography}
}

\end{document}